\documentclass[conference]{IEEEtran}
\IEEEoverridecommandlockouts
\usepackage{cite}
\usepackage{amsmath,amssymb,amsfonts}
\usepackage{algorithm}
\usepackage{algpseudocode}
\usepackage{graphicx}
\usepackage{textcomp}
\usepackage{xcolor}
\usepackage{listings}
\usepackage{url}

\lstdefinestyle{snippet}{
  basicstyle=\ttfamily\scriptsize,
  breaklines=true,
  columns=fullflexible,
  keepspaces=true,
  frame=single,
  framerule=0.3pt,
  xleftmargin=2pt,
  showstringspaces=false,
  captionpos=b,
  abovecaptionskip=4pt,
  belowcaptionskip=2pt
}

\def\BibTeX{{\rm B\kern-.05em{\sc i\kern-.025em b}\kern-.08em
    T\kern-.1667em\lower.7ex\hbox{E}\kern-.125emX}}

\begin{document}

\title{A Semantic-Layer-Mediated Agent for\\
Natural Language to SQL over Heterogeneous\\
Enterprise Databases}

\author{\IEEEauthorblockN{1\textsuperscript{st} Ha Jeong Kim}
\IEEEauthorblockA{\textit{DAQUV Corp.} \\
Seoul, Republic of Korea \\
hajkim@daquv.com}
\and
\IEEEauthorblockN{2\textsuperscript{nd} Saksonita Khoeurn}
\IEEEauthorblockA{\textit{Dept. of Management Information Systems} \\
\textit{Chungbuk National University}\\
\textit{BigDataLabs, Co., Ltd.}\\
Cheongju, Republic of Korea \\
saksonita@chungbuk.ac.kr}
\and
\IEEEauthorblockN{3\textsuperscript{rd} Ye Ji Yoon}
\IEEEauthorblockA{\textit{DAQUV Corp.} \\
Seoul, Republic of Korea \\
philoyyj@daquv.com}
\thanks{Ha Jeong Kim and Saksonita Khoeurn contributed equally to this work.}
\thanks{Corresponding author: Saksonita Khoeurn (saksonita@chungbuk.ac.kr).}
}

\maketitle

\begin{abstract}
Natural language to SQL (NL2SQL) over real enterprise databases remains
substantially harder than over academic benchmarks: schemas contain hundreds
of physical tables with opaque column names, dialects differ across engines,
and a single analytical question may require nested aggregation, time-windowed
logic, and multi-table joins. Directly prompting a large language model (LLM)
with raw schema text exposes the model to this complexity all at once and
yields brittle queries. This paper presents the architecture of a
semantic-layer-mediated NL2SQL agent that decouples \emph{intent}
from \emph{physical execution}. Rather than asking the LLM to write SQL against
raw tables, the agent reasons over a curated semantic layer through a compact
intermediate representation we call the Semantic Model Query (SMQ); a
deterministic engine compiles SMQs into dialect-correct SQL, which the agent
then inspects, composes, and executes. The system follows a single-tool
think--act loop, routes execution across SQLite, BigQuery, and Snowflake
backends, and is packaged as an end-to-end evaluation harness. Driven by
Gemini 3 Pro, the system attains 94.15\% execution accuracy on the 547-task
Spider2-snow benchmark---a suite of real-world enterprise NL2SQL tasks executed
against Snowflake---the third-highest entry on the official leaderboard and far
above schema-only baselines. We
describe the component design, the SMQ representation, the agent's exploration
strategy, and the per-backend results, and we discuss the quality and
overfitting tensions inherent to maintaining a semantic layer as a context
source for an LLM agent.
\end{abstract}


\section{Introduction}
Translating natural language questions into executable SQL (NL2SQL) is a
long-standing goal for democratizing data access. Recent large language models
(LLMs) achieve strong results on academic benchmarks such as Spider~\cite{spider}
and BIRD~\cite{bird}, where schemas are small and questions map fairly directly
to single queries. Enterprise settings are different in kind, not merely in
degree. The Spider2 benchmark~\cite{spider2} was introduced precisely to
capture this gap: its tasks operate over production data warehouses with
hundreds of columns, cryptic naming conventions, dialect-specific functions,
and questions whose gold answers routinely span dozens of lines of SQL with
common table expressions (CTEs), window functions, and multi-step aggregation.
Reported accuracies on Spider2 are far below the near-saturation numbers seen
on Spider, confirming that ``put the schema in the prompt and ask for SQL'' does
not transfer to real workloads.

We argue that the difficulty has two distinct sources. The first is
\emph{grounding}: the model must discover which physical tables and columns are
relevant and how they join, out of a large and poorly self-describing schema.
The second is \emph{composition}: the model must assemble correct, dialect-valid
SQL once the right building blocks are known. Conflating these two problems in a
single free-form generation step is what makes naive prompting brittle---a
single wrong column name or join predicate fails the whole query, and the model
has no structured surface on which to recover.

This paper presents the architecture of \emph{spider2-daquv-quvi}, an NL2SQL
agent that separates grounding from composition by interposing a \emph{semantic
layer} between the LLM and the database. The semantic layer is a curated,
business-oriented description of each database: tables are wrapped as
\emph{semantic models} exposing named \emph{dimensions}, \emph{measures}, and
\emph{metrics}, with human-readable descriptions and the physical column
expressions they map to. The agent never sees raw schema first; instead it
issues a compact, structured query---the Semantic Model Query (SMQ)---against
the semantic layer. A deterministic engine compiles each SMQ into
dialect-correct SQL and returns it. The agent uses these compiled fragments as
\emph{verified building blocks}: it inspects the physical column expressions and
join patterns they reveal, composes them into a final query (adding constructs
the SMQ compiler does not support, such as window functions or recursive CTEs),
and executes that query against the appropriate backend.

The contributions of this paper are:
\begin{itemize}
\item A system architecture for NL2SQL that mediates LLM reasoning through a
semantic layer and a structured intermediate representation (SMQ), decoupling
schema grounding from SQL composition (Sections~\ref{sec:arch}--\ref{sec:smq}).
\item A single-tool think--act agent loop in which SMQ compilation is used for
\emph{exploration} and direct SQL execution is the single terminal action,
together with the prompting strategy that enforces this discipline
(Section~\ref{sec:workflow}).
\item A multi-backend execution and evaluation harness that routes queries to
SQLite, BigQuery, and Snowflake and scores them under the Spider2 protocol
(Section~\ref{sec:exec}).
\item An empirical study on the 547-task Spider2-snow benchmark, reporting
94.15\% execution accuracy with a per-backend breakdown and a comparison
against published leaderboard methods, and a discussion of the
practical tension between semantic-layer quality and overfitting to the
evaluation set (Sections~\ref{sec:eval}--\ref{sec:discussion}).
\end{itemize}

\section{Related Work}
\textbf{Text-to-SQL benchmarks.} Spider~\cite{spider} established cross-domain,
multi-table semantic parsing as a standard task; BIRD~\cite{bird} added larger,
dirtier databases and an emphasis on efficiency and external knowledge.
Spider2~\cite{spider2} raised the bar to enterprise-scale warehouses on
BigQuery, Snowflake, and local engines, with long, realistic gold queries; it is
the benchmark we target. Our work evaluates on the Spider2-snow split.

\textbf{LLM methods for text-to-SQL.} Prompting and decomposition methods such as
DIN-SQL~\cite{dinsql}, DAIL-SQL~\cite{dailsql}, and C3~\cite{c3} improve
single-shot generation through schema linking, few-shot selection, and
self-correction. Multi-agent and pipeline systems such as
MAC-SQL~\cite{macsql} and CHESS~\cite{chess} introduce specialized
decomposer/selector/refiner roles and schema-pruning stages. These approaches
operate directly on physical schemas; our system instead routes reasoning
through a curated semantic layer, treating compiled SMQ$\rightarrow$SQL
fragments as verified grounding evidence rather than relying on the model to
recall physical names from a flat schema dump.

\textbf{Semantic layers.} Business-intelligence semantic layers and metrics
frameworks (e.g., dbt's MetricFlow~\cite{dbt}) let analysts define dimensions,
measures, and metrics once and reuse them across queries. We repurpose this idea
as an \emph{LLM grounding substrate}: the semantic layer is both the model's
view of the data and the source of compilable, dialect-correct SQL.

\textbf{LLM agents and tool use.} The think--act paradigm of ReAct~\cite{react}
interleaves reasoning traces with tool calls. Our agent adopts a strict
single-tool-per-step variant in which one tool (SMQ compilation) is for
exploration and another (SQL execution) is the single terminating action,
constraining the agent's action space to reduce error propagation.

\section{System Architecture}\label{sec:arch}
\begin{figure}[htbp]
\centerline{\includegraphics[width=\columnwidth]{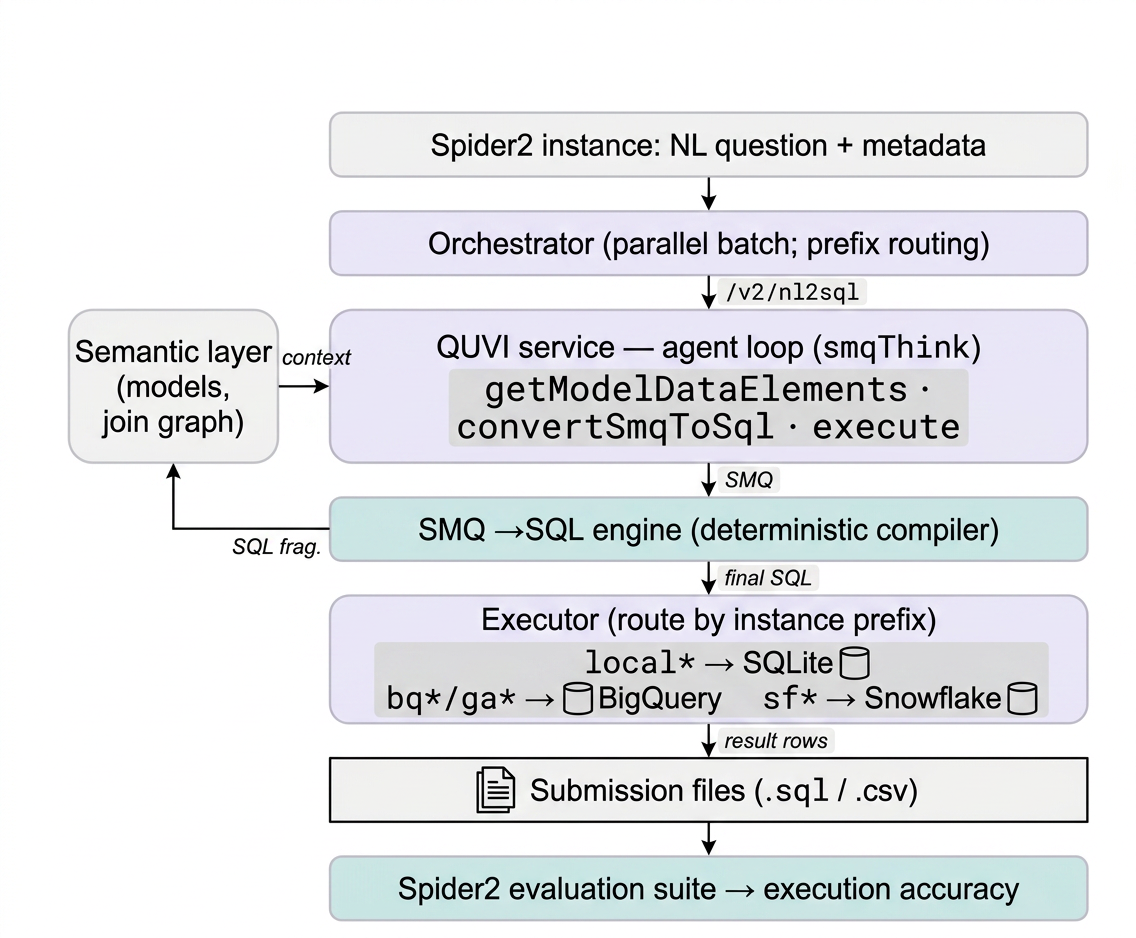}}
\caption{End-to-end architecture. The orchestrator dispatches NL questions to
the QUVI service, which runs the agent loop over the semantic layer. SMQs are
compiled to SQL by the engine; the final SQL is routed by instance prefix to
the SQLite, BigQuery, or Snowflake executor. Results are written to submission
files and scored by the Spider2 evaluation suite.}
\label{fig:arch}
\end{figure}

Fig.~\ref{fig:arch} shows the end-to-end architecture. The system is organized
into four tiers: (1) an \emph{orchestration} tier that loads benchmark instances
and drives batched, parallel execution; (2) the \emph{QUVI} NL2SQL service that
hosts the LLM agent and the semantic layer; (3) a deterministic
\emph{SMQ-to-SQL engine}; and (4) a \emph{multi-backend executor} that runs the
final SQL against the correct database. Each instance is processed as an
independent workflow keyed by an instance identifier and timestamp.

\subsection{Orchestration Tier}
The entry point parses a run specification---database split
(\texttt{lite}/\texttt{snow}), and an instance selector (explicit IDs, ranges,
prefixes, or all). An instance loader reads the Spider2 instance files, which
contain the natural-language question and metadata. An execution service then
dispatches instances through a thread pool for parallel batches, with backoff
on rate limiting (e.g., HTTP 429), per-instance timeouts, and bounded retries.
For each instance it mints a workflow identifier and calls the NL2SQL client,
which authenticates per database user before posting the question to the QUVI
service.

\subsection{QUVI NL2SQL Service}
QUVI hosts the agent and the semantic layer. On receiving a question it (i)
looks up the relevant semantic models for the target database, (ii) runs the
think--act agent loop (Section~\ref{sec:workflow}) that emits SMQs and SQL, and
(iii) delegates SMQ compilation to the engine. The semantic layer is a
per-database directory containing source table definitions, a join graph, date
and time-spine configuration, and one YAML file per semantic model.

\subsection{SMQ-to-SQL Engine}
A separate engine endpoint compiles an SMQ into SQL deterministically
(\textsf{SmqToSql}). It resolves each referenced dimension/measure/metric to its
physical column expression, injects join predicates from the join graph, and
emits SQL in the target dialect. This component is the source of \emph{ground
truth physical names}: every table name and column expression the agent uses in
its final query is lifted from a compiler output, not hallucinated.

\subsection{Multi-Backend Executor}
A custom execution server exposes a single endpoint and routes by instance-ID
prefix to the correct backend (Table~\ref{tab:routing}). Each executor holds the
appropriate credentials/driver and returns results as a list of row
dictionaries, which the orchestrator serializes to a SQL file and a CSV for
scoring.

\begin{table}[htbp]
\caption{Backend routing by instance identifier prefix}
\begin{center}
\begin{tabular}{|l|l|l|}
\hline
\textbf{Prefix} & \textbf{Backend} & \textbf{Executor} \\
\hline
\texttt{local*} & SQLite & sqlite\_executor \\
\hline
\texttt{bq*}, \texttt{ga*} & BigQuery & bigquery\_executor \\
\hline
\texttt{sf*}, \texttt{sf\_bq*} & Snowflake & snowflake\_executor \\
\hline
\end{tabular}
\label{tab:routing}
\end{center}
\end{table}

\section{The Semantic Layer and SMQ}\label{sec:smq}
\subsection{Semantic Models}
Each physical table is wrapped by a semantic model that gives it a business name
and exposes typed, described data elements. \emph{Dimensions} are aggregation
criteria (grouping/filtering keys), \emph{measures} are aggregation targets, and
\emph{metrics} are predefined aggregations or derived calculations over
measures and dimensions. Crucially, each element carries (a) a human-readable
description that the LLM reads during grounding, and (b) an \texttt{expr} field
holding the exact physical column expression used during compilation. The
abstract name and the physical expression are thus kept separate, which lets the
agent reason about \emph{intent} while the engine handles \emph{physical
mapping}. Listing~\ref{lst:semodel} shows an excerpt.

\begin{lstlisting}[style=snippet,caption={Semantic model excerpt: a dimension and a measure, each pairing a description with a physical expression.},label={lst:semodel}]
- name: RetailAnalyticsSalesModel
  table: ..._snowflake('RETAIL_ANALYTICS_SALES')
  dimensions:
  - name: asin
    type: varchar
    description: Amazon Standard Identification Number
    expr: ASIN
  measures:
  - name: orderedRevenue
    type: float
    description: ordered revenue
    expr: ORDERED_REVENUE
\end{lstlisting}

\subsection{The Join Graph}
Inter-model joins are declared once in a per-database join graph as typed edges:
a \texttt{from} model, a \texttt{to} model, a join key, and an \texttt{on}
predicate giving the left/right expressions (including transformations such as
\texttt{TRIM} on a dirty key). The compiler consults this graph to assemble
joins automatically, so the agent does not need to rediscover join predicates
for the supported cases.

\subsection{Semantic Model Query (SMQ)}
The SMQ is the intermediate representation between the agent and the engine. It
is a compact JSON object with three lists---\texttt{metrics} (query targets),
\texttt{filters} (WHERE conditions), and \texttt{group\_by}
(grouping dimensions). Elements are referenced by a uniform naming convention,
\texttt{ModelName\_\_elementName} for dimensions and measures, and by bare name
for metrics. An example SMQ and its role are shown in
Listing~\ref{lst:smq}. The SMQ deliberately covers only the common analytical
core (selection, filtering, grouping, declared joins); it does not express
CROSS JOINs, arbitrary subqueries, advanced window functions, or recursive
CTEs. This is by design: the SMQ is an \emph{exploration and grounding}
instrument, and the long tail of SQL complexity is handled by the agent
composing on top of compiler outputs.

\begin{lstlisting}[style=snippet,caption={An SMQ. The engine compiles it to dialect-correct SQL and returns the SQL plus a result preview.},label={lst:smq}]
{
  "metrics": ["RetailAnalyticsSalesModel__orderedRevenue"],
  "filters": ["RetailAnalyticsSalesModel__period = 'DAILY'"],
  "group_by": ["RetailAnalyticsSalesModel__asin"]
}
\end{lstlisting}

\section{Agent Workflow}\label{sec:workflow}
\subsection{Think--Act Loop}
The agent runs a constrained think--act loop. Each turn it emits a private
reasoning block followed by exactly one tool call. Three tools are available:
\texttt{getModelDataElements} (list the metrics/dimensions of selected models),
\texttt{convertSmqToSql} (compile an SMQ, returning the SQL and a five-row
result preview), and \texttt{execute} (run a final SQL query). Restricting each
response to a single tool call narrows the action space and makes the trajectory
auditable. Algorithm~\ref{alg:loop} summarizes the loop.

\begin{algorithm}[htbp]
\caption{Semantic-layer-mediated NL2SQL agent loop}
\label{alg:loop}
\begin{algorithmic}[1]
\State \textbf{Input:} question $q$, semantic models $M$ for target DB, dialect $d$
\State $E \gets \textsc{getModelDataElements}(\text{relevant}(M, q))$
\Comment{discover elements}
\State $B \gets \emptyset$ \Comment{verified SQL building blocks}
\Repeat
  \State write an SMQ $s$ for an intermediate result
  \State $(sql, preview) \gets \textsc{convertSmqToSql}(s)$
  \State $B \gets B \cup \{(\text{expr/table names from } sql)\}$
\Until{building blocks suffice for $q$}
\State compose final SQL $Q$ from $B$ (add CTEs, window fns, subqueries)
\State \textbf{return} $\textsc{execute}(Q)$ \Comment{single terminal action}
\end{algorithmic}
\end{algorithm}

\subsection{SMQ-for-Exploration Discipline}
The defining prompt strategy is that \texttt{convertSmqToSql} is used for
\emph{exploration}, not for producing the final answer. Its purpose is to reveal
how abstract elements map to physical columns, the dialect's syntax, and join
patterns. The agent extracts physical table names and column expressions from
the compiled SQL and assembles its own final query, applying constructs beyond
the SMQ's expressiveness. A strict constraint forbids referencing semantic model
names as physical tables in any executed SQL: only names lifted from a compiler
output may appear. Direct schema introspection (e.g., querying
\texttt{INFORMATION\_SCHEMA}) is disallowed, forcing all grounding through the
semantic layer. The single \texttt{execute} call is terminal: a successful
execution ends the workflow.

\subsection{Why Mediation Helps}
This design addresses both difficulty sources from
Section~\ref{sec:arch}. Grounding becomes a guided search over described,
business-named elements rather than a guess over a flat schema; and every
physical identifier the model commits to has already been validated by the
compiler, so composition errors are confined to the structure the agent adds on
top, where they are easier to detect and repair via re-exploration.

\section{Execution and Evaluation Harness}\label{sec:exec}
Final SQL is dispatched to the backend executor and the returned rows are
written as a per-instance SQL file and CSV under a submission directory, with a
human-readable Markdown trace of the full tool-call history saved alongside for
debugging. Scoring uses the official Spider2 evaluation suite in two modes:
execution-result comparison (the submitted result set is compared against the
gold result) and SQL comparison. Configuration (service ports, data and
semantic-layer paths, timeouts, retry limits, credentials) is centralized in a
single application config so the same harness runs across the SQLite, BigQuery,
and Snowflake environments without code changes.

\section{Evaluation}\label{sec:eval}
\subsection{Setup}
We evaluate on Spider2-snow, a benchmark of 547 enterprise NL2SQL instances
executed against Snowflake, spanning databases of SQLite-, BigQuery-, GA4-, and
native-Snowflake origin. The
underlying LLM is Gemini 3 Pro (preview), invoked through the semantic-layer
\texttt{smqThink} agent node at temperature 0.1 with extended thinking enabled
at the high setting, a 16{,}384-token output budget, and a cap of 20 SMQ
iterations per instance; the SMQ$\rightarrow$SQL compiler is deterministic. We
find the agent's accuracy to be sensitive to the model's thinking
configuration: extended thinking is what allows the model to compose the long,
multi-step SQL that Spider2-snow demands, and disabling it substantially
degrades accuracy. We report execution accuracy: an instance is correct if the
executed query's result set matches the gold result under the official Spider2
execution-result protocol.

\subsection{Overall Result}
On Spider2-snow the system answers 515 of 547 instances correctly, an execution
accuracy of \textbf{94.15\%}. Given that Spider2 gold queries are long,
multi-step, and dialect-specific, this result indicates that semantic-layer
mediation, paired with a strong reasoning model and well-curated per-database
semantic layers, provides highly effective grounding on real enterprise schemas.

\subsection{Per-Backend Breakdown}
Table~\ref{tab:results-snow} breaks accuracy down by instance class. Accuracy
exceeds 94\% on every database-origin class except native Snowflake
(\texttt{sf}, 55.6\%), which is also the smallest class (18 instances) and thus
the highest-variance. Backend-specific dialect handling and the maturity of each
database's semantic layer---not the agent loop itself---are the dominant factors
in the remaining failures.

\begin{table}[htbp]
\caption{Execution accuracy on Spider2-snow by database-origin class (Gemini 3 Pro)}
\begin{center}
\begin{tabular}{|l|c|c|}
\hline
\textbf{Instance class (origin)} & \textbf{Correct/Total} & \textbf{Accuracy} \\
\hline
\texttt{sf\_local} (SQLite-origin) & 131 / 135 & 97.0\% \\
\hline
\texttt{sf\_ga} (GA4-origin) & 24 / 25 & 96.0\% \\
\hline
\texttt{sf\_bq} (BigQuery-origin) & 350 / 369 & 94.9\% \\
\hline
\texttt{sf} (native Snowflake) & 10 / 18 & 55.6\% \\
\hline
\textbf{Total} & \textbf{515 / 547} & \textbf{94.15\%} \\
\hline
\end{tabular}
\label{tab:results-snow}
\end{center}
\end{table}

\subsection{Comparison with Published Methods}
Table~\ref{tab:compare} places our Spider2-snow result against entries on the
official Spider2 leaderboard~\cite{spider2lb}. General agent and prompting
pipelines on raw schemas remain low on this benchmark---DAIL-SQL~\cite{dailsql}
with GPT-4o reaches only 2.2\%, Spider-Agent~\cite{spider2} 23--26\% depending
on the backbone, and ReFoRCE~\cite{reforce} 31.3\%. Our semantic-layer-mediated
agent (listed as \emph{QUVI-3 + Gemini-3-pro-preview}) attains \textbf{94.15\%},
the third-highest entry overall. The large margin over schema-only baselines is
consistent with our central claim that mediating the LLM through a curated
semantic layer is decisive for enterprise NL2SQL; we note, however, that the
semantic layer encodes per-database domain knowledge that the zero-shot
baselines do not have access to, so the comparison reflects the value of the
\emph{system} (semantic layer plus agent), not the LLM alone.

\begin{table}[htbp]
\caption{Execution accuracy on Spider2-snow: published methods (Spider2 leaderboard)}
\begin{center}
\begin{tabular}{|l|c|}
\hline
\textbf{Method} & \textbf{Accuracy} \\
\hline
DAIL-SQL + GPT-4o~\cite{dailsql} & 2.20\% \\
\hline
Spider-Agent + o1-preview~\cite{spider2} & 23.58\% \\
\hline
Spider-Agent + Claude-4-Sonnet~\cite{spider2} & 25.78\% \\
\hline
ReFoRCE + o1-preview~\cite{reforce} & 31.26\% \\
\hline
\textbf{Ours (QUVI-3 + Gemini 3 Pro)} & \textbf{94.15\%} \\
\hline
\end{tabular}
\label{tab:compare}
\end{center}
\end{table}

\subsection{Failure Modes}
From the saved execution traces, recurring failure modes include: (i) SMQ
compilation gaps, where a referenced model lacks a needed join relation and the
compiler cannot assemble the fragment; (ii) composition errors in the
agent-authored portion of complex queries (nested time windows, exact rounding
and ranking semantics); and (iii) under-specified semantic layers, where a
column needed by the gold query is not yet exposed as a described element. The
first and third are addressed by improving the semantic layer rather than the
agent, which motivates the discussion below.

\section{Discussion}\label{sec:discussion}
\subsection{Semantic-Layer Quality is the Lever}
Because grounding is delegated to the semantic layer, system accuracy is
strongly bounded by how well each database's models describe the data and
declare its joins. In practice, improving a database's semantic models---adding
missing elements, sharpening descriptions, declaring join edges---directly
lifts accuracy on that database. This makes the semantic layer the primary
maintenance surface.

\subsection{The Overfitting Tension}
Curating the semantic layer against an evaluation set introduces a Goodhart-style
risk: descriptions can drift from concise, generalizable annotations toward
verbose text that effectively encodes the expected answer for known questions.
Such context may raise benchmark accuracy while reducing robustness to unseen
questions and schema change---the analogue of reward-proxy overoptimization. We
therefore treat the semantic layer as code subject to review: descriptions
should encode reusable schema knowledge, not question-specific hints, and
structural information (joins, expressions) should live in typed fields rather
than prose. Quantifying and regularizing this trade-off---an intrinsic quality
signal for constructed context that is independent of downstream task
accuracy---is an open problem and a direction for future work.

\subsection{Limitations}
The SMQ deliberately covers only a common analytical core, so the hardest
queries depend on agent-authored SQL whose correctness the engine does not
guarantee. The evaluation is single-benchmark (Spider2-snow) and uses
execution-result matching, which can over- or under-credit edge cases.
Per-backend results for the smallest classes are high-variance. Finally,
accuracy is entangled with the maturity of each database's hand-authored
semantic layer, which varies across databases.

\section{Conclusion}
We presented the architecture of a semantic-layer-mediated NL2SQL agent that
decouples schema grounding from SQL composition. By interposing a curated
semantic layer and a compact SMQ intermediate representation between the LLM and
the database, and by using deterministic SMQ$\rightarrow$SQL compilation as a
source of verified building blocks within a constrained single-tool agent loop,
the system reaches 94.15\% execution accuracy on the 547-task Spider2-snow
benchmark---the third-highest entry on the official leaderboard and far above
schema-only baselines. The results indicate that
mediating LLM reasoning through a structured, business-oriented layer is an
effective strategy for enterprise NL2SQL, and that the central remaining
challenge is maintaining semantic-layer quality without overfitting it to the
evaluation set.

\end{document}